\documentclass[10pt,twocolumn,letterpaper]{article}

\usepackage[pagenumbers]{cvpr} 

\usepackage{graphicx}
\usepackage{amsmath}
\usepackage{amssymb}
\usepackage{booktabs}
\usepackage[table,xcdraw]{xcolor}

\usepackage[pagebackref,breaklinks,colorlinks]{hyperref}

\usepackage[capitalize]{cleveref}
\crefname{section}{Sec.}{Secs.}
\Crefname{section}{Section}{Sections}
\Crefname{table}{Table}{Tables}
\crefname{table}{Tab.}{Tabs.}

\def\cvprPaperID{3609} 
\def\confName{CVPR}
\def\confYear{2023}

\begin{document}

\title{Object Preserving Siamese Network for Single Object Tracking on Point Clouds}

\author{Kaijie Zhao, Haitao Zhao, Zhongze Wang, Jingchao Peng, Zhengwei Hu\\
East China University of Science and Technology\\
No.130, Meilong Rd, Shanghai\\
}
\maketitle

\begin{abstract}
   Obviously, the object is the key factor of the 3D single object tracking (SOT) task. However, previous Siamese-based trackers overlook the negative effects brought by randomly dropped object points during backbone sampling, which hinder trackers to predict accurate bounding boxes (BBoxes). Exploring an approach that seeks to maximize the preservation of object points and their object-aware features is of particular significance. Motivated by this, we propose an Object Preserving Siamese Network (\textbf{OPSNet}), which can significantly maintain object integrity and boost tracking performance. Firstly, the \textbf{object highlighting module} enhances the object-aware features and extracts discriminative features from template and search area. Then, the \textbf{object-preserved sampling} selects object candidates to obtain object-preserved search area seeds and drop the background points that contribute less to tracking. Finally, the \textbf{object localization network} precisely locates 3D BBoxes based on the object-preserved search area seeds. Extensive experiments demonstrate our method outperforms the state-of-the-art performance ($\sim$9.4\% and $\sim$2.5\% success gain on KITTI and Waymo Open Dataset respectively).
\end{abstract}

\section{Introduction}
\label{sec:intro}

Single object tracking (SOT) has become a significant issue of computer vision, which contributes widely to various fields, such as autonomous driving \cite{liu2020deep,luo2018fast}, security surveillance \cite{xing2010multiple,tang2017multiple} and robotics \cite{budiharto2020design,jiang2021high}, etc. With the fast development of 3D sensors, such as LiDAR, 3D data reflecting real-world coordinates and object sizes can be captured. Therefore, 3D computer vision tasks attract more and more attention in recent years \cite{ioannidou2017deep,lu2020survey}. 3D single object tracking is also an essential part of 3D computer vision, which aims to improve autonomous cars' safety via predicting accurate object trajectories. At present, many 2D single object tracking algorithms have been proposed and proved their efficiency, but they cannot directly handle 3D data since 3D point clouds are of sparsity and non-uniformity \cite{qi2020p2b}.

\begin{figure}
    \centering
    \includegraphics[scale=0.40]{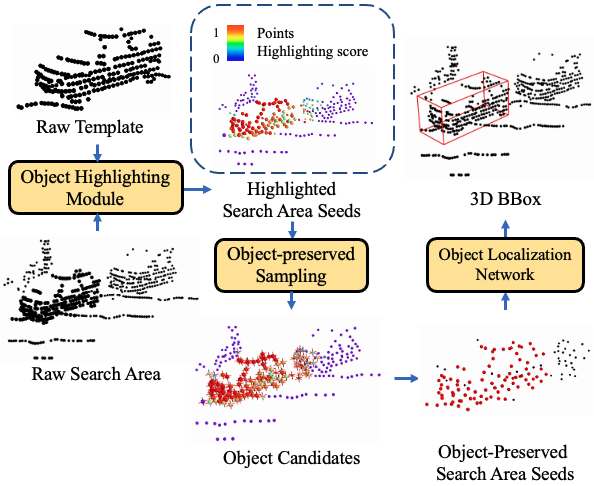}
    \caption{Exemplified illustration to show how our OPSNet works, from the object highlighting module to the object-preserved sampling and localization module.}
    \label{fig:exemplify}
\end{figure}
\noindent

Mainstream 3D SOT approaches are based on the Siamese network \cite{giancola2019leveraging,qi2020p2b,zheng2021box,zhou2022pttr,hui20213d,hui20223d} that consists of two weight-shared backbone branches to respectively handle template and search area. Most of the Siamese-based approaches use PointNet++ \cite{qi2017pointnet++} as the backbone which hierarchically samples points and aggregates local features from raw point clouds. STNet \cite{hui20223d} firstly adopts Transformer as the backbone for encoding all the points of template and search area. Considering that the tracking tasks require to be conducted in real-time and tracking speed is also a significant evaluation indicator, the PointNet++ is of higher efficiency and less computational burden \cite{qi2017pointnet++}.

Since PointNet++ applies furthest points sampling (FPS) or random sampling, all the approaches based on PointNet++ suffer from the negative effects brought by randomly dropped object points. To overcome this issue, PTTR \cite{zhou2022pttr} samples the search area points that are most similar to template points in the embedding space for object preservation. However, once PTTR regresses an inaccurate 3D BBox as a template during inference, search area sampling will be simultaneously misled. Exploring a robust approach that seeks to maximize the preservation of object points and their object-aware features for tracking is of particular significance. We propose our novel OPSNet for object highlighting, object preservation, and accurate object localization. 

We exemplify how our OPSNet works in \cref{fig:exemplify}. The object highlighting module consists of three main parts: 1) The backbone first obtains template and search area seeds (each seed is represented as $[x;f]\in \mathbb{R}^{3+d_1}$, where $x$ denotes the 3D coordinates of the seed and $f$ is the seed features vector), then feature-targeted transformation converts the search area seeds into two feature-targeted seed subsets, 2) adaptive cross-correlation utilizes two branches to respectively handle two feature-targeted subsets, one branch highlights object-aware features from the consistency between the subset and the template; the other branch extracts discriminative features from the discrepancy between the subset and the template, 3) object augmentation concatenates the two subsets and further obtains highlighted search area seeds with the cross attention mechanism. Each seed is appended with a highlighting score predicted by MLP,  the score is supervised by smooth point-wise one-hot label and Focal Loss \cite{lin2017focal}. 

Obtained the highlighted search area seeds and the highlighting scores, the object-preserved sampling selects the seeds with top $k$ highlighting scores as object candidates, Then, each object candidate clusters the neighbor seeds to aggregate local features, and we can obtain the object-preserved search area seeds. Object integrity and object-aware features can be effectively preserved and the background points that cause redundancy have been dropped simultaneously. Finally, the object localization network performs effective BBoxes prediction based on the object-preserved search area seeds without generating multiple proposals. 

Experimental results (average success gain of $\sim$9.4\% and $\sim$2.5\% over the SOTA method on KITTI \cite{geiger2012KITTI} and Waymo Open Dataset \cite{sun2020scalability}) demonstrate that OPSNet achieves high-performance tracking, handles sparse point clouds effectively, and performs robust tracking on long sequences.

The contributions of this paper are as follows: 

\begin{itemize}

\item Object highlighting module adaptively extracts not only highlighted object-aware features but also discriminative features from template and search area. 
\item Object-preserved sampling significantly maintains the object integrity and reduces the misleading background points and redundancy.
\item Object localization network performs more effective BBoxes prediction than other voting-based regression approaches without generating multiple proposals.
\end{itemize}

\section{Related Work}
\noindent \textbf{Deep learning on point clouds.} Since Qi \textit{et al.} \cite{qi2017pointnet} proposed their inspirational paradigm PointNet, 3D deep learning on point clouds has stimulated the interest of researchers. 3D deep learning on point clouds methods mainly consist of point-based \cite{ding2019votenet,yang20203dssd,qi2020p2b,qi2017pointnet++}, voxel-based \cite{zhou2018voxelnet,zhuang2022span,lang2019pointpillars}, graph-based \cite{wang2019dynamic,natali2011graph}. We choose the point-based method since the point-based method learns distinct object-aware features, handles sparse scenes well and reduces computational burden \cite{qi2017pointnet++}.

\noindent \textbf{3D Single object tracking.} Recently, with the rapid development of sensors, 3D sensors such as LiDAR can obtain 3D point cloud data for 3D single object tracking. Giancola \textit{et al.} \cite{giancola2019leveraging} first proposed a shape-completion-based 3D Siamese tracker (SC3D), which originally encodes shape information into the template for matching template and search area. However, SC3D cannot be end-to-end trained.  Qi \textit{et al.} \cite{qi2020p2b} proposed a point-to-box (P2B) Siamese tracking scheme, which adopted the target-specific feature augmentation module to learn template and search area cross-correlation and utilized Hough Voting modules to regress potential target centers in the search area. Zhou \textit{et al.} \cite{zhou2022pttr} proposed PTTR that firstly adopted relation-aware sampling (RAS) in the Siamese network, search area branch obtains a feature map shared by the template for sampling reference, which effectively preserves object points in the search area. Furthermore, Hui \textit{et al.} \cite{hui20213d} proposed a novel BBox regression module, which compressed voxelized sparse 3D features for 2D detection and performs effective object tracking. 

\begin{figure*}
    \centering
    \includegraphics[scale=0.40]{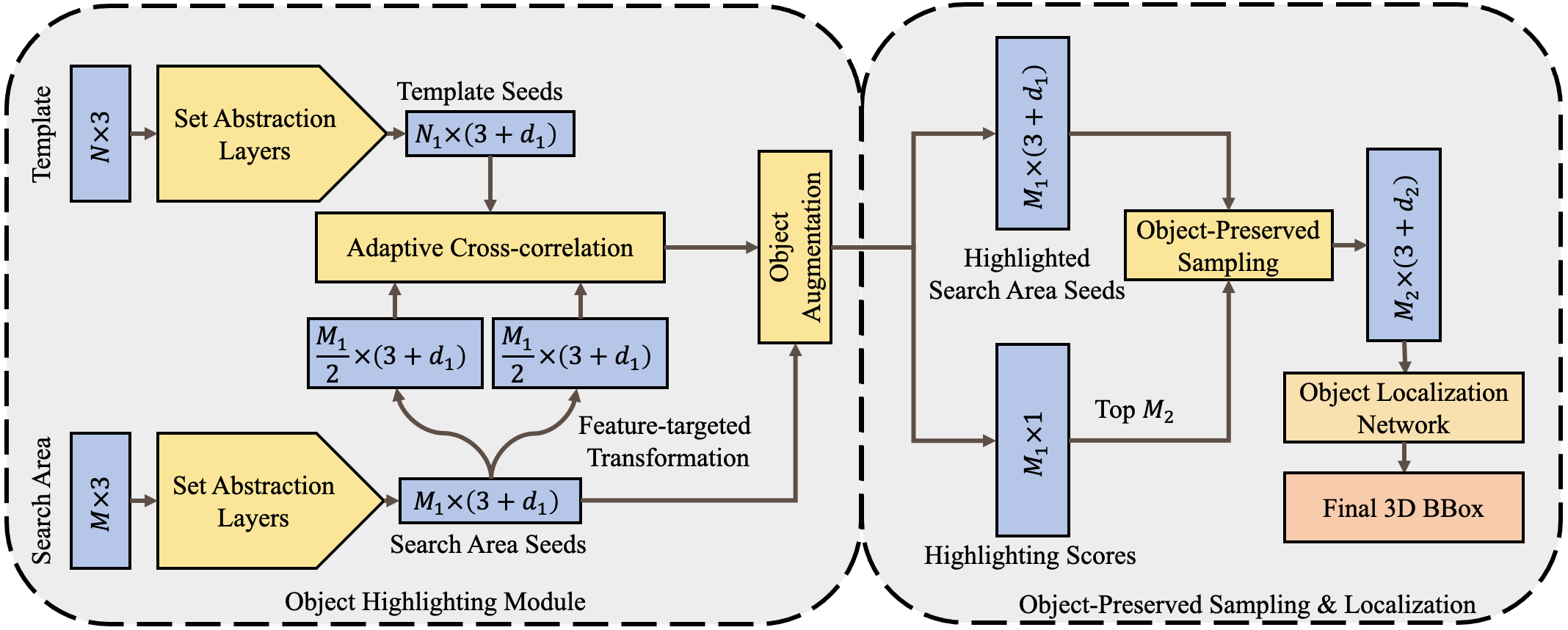}
        \caption{The main pipeline of OPSNet. OPSNet consists of two main parts: 1) object highlighting module, and 2) object-preserved sampling and localization. Firstly, the set abstraction layers output template and search area seeds, respectively. Then, with the help of template seeds object highlighting module enhances the object-aware features and extracts the discriminative features from search area seeds. Finally, the object-preserved sampling selects object candidates from the highlighted search area seeds for object preservation and accurate 3D BBox localization.}
    \label{fig:mainpipe}
\end{figure*}

\noindent \textbf{Object segmentation.} Object-preserved sampling employed in OPSNet is similar to object segmentation, but they differ in task-driven purposes. Recently proposed object segmentation approaches \cite{brodeur2021point,ma2019multi,luo2021boundary,song2022ogc} are based on deep learning. Wu \textit{et al.} \cite{squeezev1} formulated object segmentation as a point-wise multi-class classification problem, and they propose an end-to-end pipeline called SqueezeSeg based on convolutional neural networks. Our object-preserved sampling also aims at point-wise classification, but the task-driven purpose is selecting object candidates by referring to the point-wise highlighting scores.

\section{Method} 
\subsection{Overview}
In Siamese-based tracking methods, the backbone downsamples template and search area to obtain their seeds by set abstraction layers that consist of sampling and grouping \cite{qi2017pointnet++}, each seed is represented as $[x;f]\in \mathbb{R}^{3+d_1}$, where $x$ denotes the 3D coordinates of the seed and $f$ is the seed features vector. We aim to retain object points by object-preserved sampling and highlight object-aware features in the search area for accurate object localization. OPSNet has two main parts (\cref{fig:mainpipe}): 1) object highlighting module, and 2) object-preserved sampling and localization. The object highlighting module aims to enhance object-aware features and extract discriminative features from template and search area, which can assist the object-preserved sampling to achieve object-background recognition. Object-preserved sampling and localization maintain object integrity and reduce redundancy for effectively predicting accurate BBoxes.

\subsection{Object Highlighting Module}
Raw points in template $P_{tmp}$ (of size $N$) and search area $P_{sea}$ (of size $M$) are fed into a feature backbone to obtain $N_1$ template seeds $Q=\{q_i\}^{N_1}_{i=1}$ and search area seeds $R=\{r_j\}^{M_1}_{j=1}$, the backbone consists of two set abstraction layers. Then feature-targeted transformation module converts the search area seeds into two feature-targeted seed subsets. After that, the adaptive cross-correlation utilizes two branches to respectively handle two subsets, one branch highlights object-aware features with consistent cross-correlation; the other branch extracts discriminative features with discrepant cross-correlation. Finally, the object augmentation module further enhances the object-aware features.  

\noindent \textbf{Feature-targeted transformation.}
Feature-targeted transformation initializes two independent MLPs with input and output dimension $\{M_1, \frac{M_1}{2}\}$, then the transposed search area seeds $R^{\rm{T}}$ are fed into them to generate two feature-targeted seed subsets $R^{\rm{T}}_1$ and $R^{\rm{T}}_2$, respectively. For dimension alignment, we transpose $R^{\rm{T}}_1$ and $R^{\rm{T}}_2$ to obtain feature-targeted subsets $R_1$ (of size $\frac{M_1}{2}$) and $R_2$ (of size $\frac{M_1}{2}$).

\noindent \textbf{Adaptive Cross-correlation.}
Adaptive cross-correlation consists of consistent and discrepant cross-correlation. 

\begin{figure*}
    \includegraphics[scale=0.39]{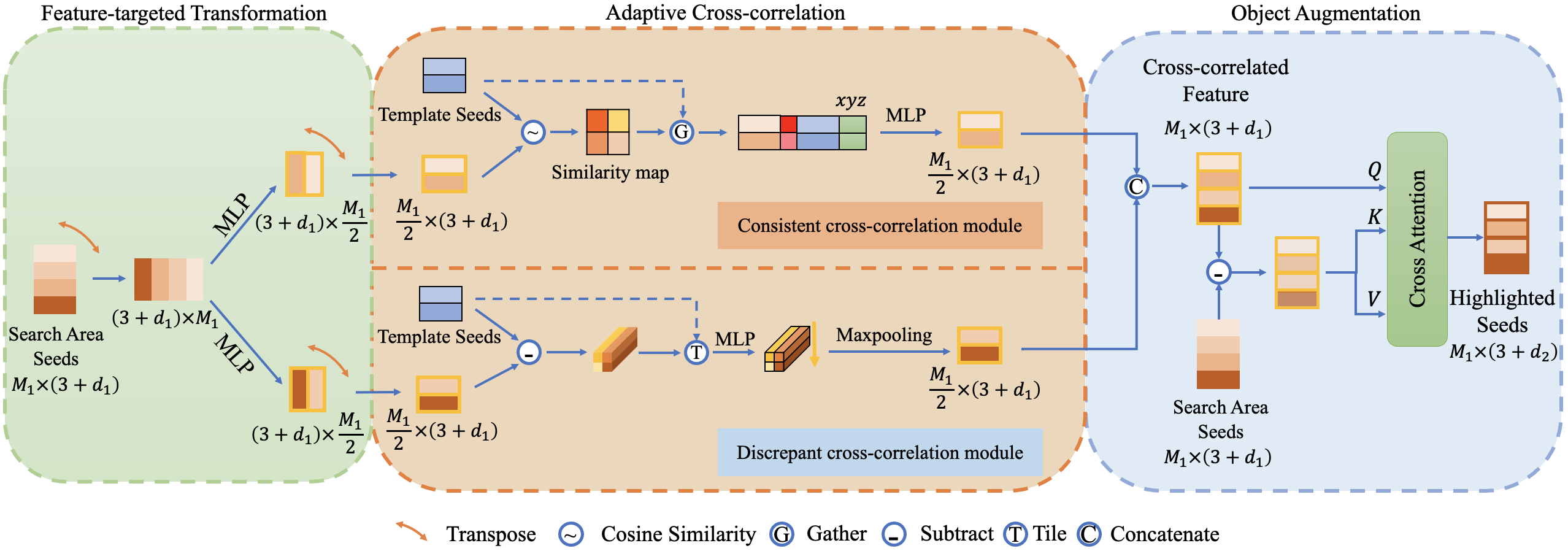}
    \caption{Illustration of the object highlighting module. Object highlighting module can be divided into three main parts: 1) feature-targeted transformation converts search area seeds into two feature-targeted seed subsets, 2) adaptive cross-correlation extracts the consistent and discrepant features into two subsets respectively with the help of the template seeds, 3) object augmentation further enhances the object features by employing the cross attention mechanism.}
    \label{fig:con_dis}
\end{figure*}
\noindent \textit{Consistent cross-correlation.} Consistent cross-correlation computes the cosine similarity $F_{cos}$ between a seed subset ${R_1}$ and the template seeds $Q$, which focuses more on consistent semantic correlation. 
\begin{flalign}
    s_{ij}=F_{{\rm cos}} (q_i,r_j)=\frac{{q_i}^{\rm{T}} \cdot {r_j}}{{\Vert q_i \Vert}_2 \cdot {\Vert r_j \Vert}_2}
\end{flalign}
where ${q_i,r_j}$ respectively denotes certain point in template seeds $Q$ and feature-targeted subset $R_1$. $s_{ij}$ denotes the cosine similarity between points $q_i,r_j$. Then we gather the most similar point in template seeds with each point in the subset $R_1$, as \cref{fig:con_dis} shows, which can be written as :
\begin{flalign}
    \begin{split}
        &{r_j}'={\rm MLP}[r_j,s_{kj},q_k,x_k] \\
        &k=\mathop{{\rm argmax}}\limits_{i=1,2,...,N_1} \left\{ F_{{\rm cos}}(q_i,r_j) \right\}, {\forall}r_j \in R_1
    \end{split}
\end{flalign}

where $k$ is the index of the maximum similarity $F_{\rm cos}(q_i,r_j)$  value, and $s_{kj},x_k$ are the corresponding maximum value and coordinate $(x,y,z)$, respectively. $[\cdot,\cdot,\cdot,\cdot]$ denotes concatenation. 

Here, we obtain a consistent cross-correlated feature vector ${r_j}'$ of $j$-th point in the subset $R_1$ after MLP and integrate them to generate consistent cross-correlated seeds subset ${R_1}'$.

\noindent \textit{Discrepant cross-correlation.} Discrepant cross-correlation extracts discrepant semantics between $Q$ and feature-targeted subset $R_2$ into $R_2$. We firstly learn a discrepancy map by computing subtracted difference and using an MLP, which is formulated as:
\begin{flalign}
    d_{ij}=F_{{\rm sub}}(q_i,r_j)={\rm MLP}(q_i-r_j),\forall q_i \in Q, r_j \in R_2
\end{flalign}
where $q_i,r_j$ respectively denote certain point in the template seeds $Q$ and subset $R_2$. $d_{ij}$ denotes the discrepant weight between points $q_i,r_j$, as \cref{fig:con_dis} shows. The discrepant cross-correlated targeted point ${r_j}''$ is given by:
\begin{equation}
    {r_j}''={\rm MLP}(\mathop{{\rm MAX}}\limits_{i=1,2,...,N_1} \left\{ q_i \cdot d_{ij} \right\})
\end{equation}
here MAX is the max pooling operator. We obtain a discrepant cross-correlated feature vector ${r_j}''$ of $j$-th point in the subset $R_2$ and integrate them to generate discrepant cross-correlated seeds subset ${R_2}'$.

Concatenating ${R_1}'$ and ${R_2}'$, the cross-correlated search area seeds $R'$ are obtained. 
 
\noindent \textbf{Object augmentation.} We first subtract the cross-correlated seeds $R'$ and search area seeds $R$, then put the subtracted tensor into MLP for augmenting differentiation features ${R^\Delta}$. For further augmenting the object-aware features in the cross-correlated seeds $R'$, cross attention mechanism is adopted, which can be formulated as:
\begin{flalign}
\widetilde{R}={\rm CrossAttention}(R'+X^s, R^\Delta+X^s, R^\Delta+X^s) 
\label{crossatt}
\end{flalign}
where the three inputs from left to right are used as query, key, and value in \cref{crossatt}, respectively. $X^s$ denotes the position embedding.

Here the object-highlighted search area seeds $\widetilde{R}=\{f_j\}^{M_1}_{j=1}$ are obtained.

\subsection{Object Preserved Sampling} Obtained object-highlighted search area seeds $\widetilde{R}=\{f_j\}^{M_1}_{j=1}$ and given original BBox annotations, point-wise one-hot labels $Y=\{y_i\}^{M_1}_{i=1}$ can be generated. We predict a highlighting score for each seed: $\hat{y_i}={\rm MLP} (f_i),i=1,2,...,M_1$, and select seeds with highest $M_2$ scores as candidates. In addition, smooth one-hot labels is formulated as:
\begin{equation}
{{y_i}'}=
\left\{
             \begin{array}{lr}
             y_i-\frac{\varepsilon}{M_1}, &  y_i=1.\\
             y_i+\frac{\varepsilon}{M_1}, & y_i=0. \label{labelsmooth}
             \end{array}
\right.
\end{equation}

Obtained the smooth one-hot labels, we compute the Focal loss \cite{lin2017focal} for supervision. Considering the search area generally contains unbalanced objects and background points number, the focal loss can be regarded as an ingenious solution, which is defined by:
\begin{flalign}
    \begin{split}
        &\mathcal{L}_{cls}=-\sum {{y_i}'}[y_i=1] {(1-\hat{y_i})^\alpha} {\rm log}(\hat{y_i})+ \\
        &{{y_i}'}[y_i=0] {\hat{y_i}^\alpha} {\rm log}(1-\hat{y_i}) 
    \end{split}
\end{flalign}

where ${{y_i}'}[cond.]$ is the indicator function, follows Eq. (\ref{labelsmooth}), $\hat{y_i} $ is the predicted highlighting score. Besides, $\alpha=2$ is empirically set in all experiments. 

We select object candidates with top $M_2$ highlighting scores as ball query centers for grouping and aggregating local features to obtain object-preserved search area seeds $R^{op}=\{r^{op}_j\}^{M_2}_{j=1}$, $r^{op}_j=[x^{op}_j;f^{op}_j] \in \mathbb{R}^{3+d_2}$, where $x^{op}$ denotes the object-preserved seed coordinates and $f^{op}$ is the feature vector.

\subsection{Object Localization Network}
Inspired by \cite{ge2020afdet,hui20213d}, we develop an object localization network utilizing bird's eyes view (BEV), and the object detection paradigm has been customized for tracking adaption.

Firstly, sparse search area seeds $R^{op}$ requires to be segmented within the BEV range $[(x_{min},x_{max}),(y_{min},y_{max})]$. We voxelize segmented $R^{op}$ with voxel side length as $r$,  then we can obtain dense BEV feature map $\mathcal{E} \in \mathbb{R}^{H \times W \times c}$ in $x$-$y$ plane, where $H=\lfloor \frac{x_{max}-x_{min}}{r} \rfloor, W=\lfloor \frac{y_{max}-y_{min}}{r} \rfloor$, and $c$ is the channel of features that obtained by average pooling. 
Then we locate the original 3D point coordinates $(x,y,z)$ in 2D $x$-$y$ plane presented by $(u,v)$, where $u=\frac{x-x_{min}}{r}$ and $v=\frac{y-y_{min}}{r}$. 
Since the dense map uses discrete coordinates, we need discrete $(\Bar{u},\Bar{v})=(\lfloor u \rfloor, \lfloor v \rfloor)$ as well. Here $\lfloor \cdot \rfloor$ denotes floor operation.

Therefore, we can localize 2D-center based on the dense BEV feature map, we discretize ground truth bounding box (GTBB) center $\widetilde{p^c}$ $(x,y,z)$ into $p^c$ $(\Bar{u},\Bar{v})$, then we expand $p^c$ into a center map $\mathcal{M} \in \mathbb{R}^{H \times W \times 1}$ that defined by:
\begin{flalign}
    \left\{
             \begin{array}{lr}
             p_{ij}=\frac{1}{d+1}, &  p_{ij} \in \textbf{B}.\\
             p_{ij}=0, & p_{ij} \notin \textbf{B}.
             \end{array}
    \right.
\end{flalign}
where $d$ presents the Euclidean distance between $p_{ij}$ and GTBB center $p^c$; $p_{ij} \in \textbf{B}$ and $p_{ij} \notin \textbf{B}$ mean the pixel $p_{ij}$ inside or outside the 2D GTBB $\textbf{B}$, respectively. 

Pervasive distractors can easily mislead object center localization, and we need to localize the only object required to track with, thus we conduct train \& test syncretic loss punishment on wrongly located centers. We firstly predict $n$ stacked center maps $\hat{\mathcal{M}_n} \in \mathbb{R}^{H \times W \times n}$ from BEV feature map ${\mathcal{E}}$ but we only enforce the first center map $\hat{\mathcal{M}_0} \in \mathbb{R}^{H \times W \times 1}$ to approach the ground truth ${\mathcal{M}}$ by using Focal loss \cite{lin2017focal}, which denoted by $\mathcal{L}_{center}$. As for other $n-1$ center maps, we compute Euclidean distances summation between the predicted centers $\hat{p^c_i}$ in $\hat{\mathcal{M}_i},(i=1,2,...,n)$ and GT, which is defined by:
\begin{flalign}
    \mathcal{L}_{loc}=\frac{1}{n}\sum^k_{i=1}{\Vert p^c - \hat{p^c_i} \Vert_2}
\end{flalign}
where $p^c_i$ denotes $i$-th predicted center in $\hat{\mathcal{M}_i}$.

Since the 2D BEV-based feature map generates a discrete 2D-center, the offset of the continuous ground truth center requires to be regressed. We surround the predicted object center with a square of radius $l$. Similar to 2D-center prediction, predicted offset map $\hat{\mathcal{O}} \in \mathbb{R}^{H \times W \times 2}$ is generated from BEV feature map ${\mathcal{E}}$, which is supervised by: 
\begin{flalign}
    \mathcal{L}_{off}=\sum_{\delta{x}=-l}^l{\sum_{\delta{y}=-l}^l{\lvert {\hat{\mathcal{O}}_{p^c+(\delta{x},\delta{y})}-[\widetilde{p^c}-p^c+(\delta{x},\delta{y})]} \rvert}}
\end{flalign}
where $\widetilde{p^c}$ and $p^c$ present the ground truth center with continuous and discrete coordinates respectively.

Two individual CNNs are applied to directly regress the $z$-axis coordinate and the rotation angle $\theta$. Given two predicted z-axis map $\hat{\mathcal{Z}} \in \mathbb{R}^{H \times W \times 1}$ and the rotation map $\hat{\Theta} \in \mathbb{R}^{H \times W \times 1}$, we use $L_1$ loss to compute error:
\begin{flalign}
    \mathcal{L}_{z}=\lvert \hat{\mathcal{Z}_c} - z \rvert \\
    \mathcal{L}_{\theta}=\lvert \hat{\Theta} - \theta \rvert
\end{flalign}

We sum the all above losses as our final network loss: $\mathcal{L}=\lambda_1(\mathcal{L}_{center}+\mathcal{L}_{loc}+\mathcal{L}_{off}+\mathcal{L}_{\theta})+\lambda_2\mathcal{L}_{z}+\lambda_3\mathcal{L}_{cls}$. $\lambda_1$, $\lambda_2$, and $\lambda_3$ are the hyper-parameters for loss regulation.

\section{Experiments} \label{sec:exps}
Extensive experiments have been conducted to validate the effectiveness of our OPSNet and its components, we mainly focus on car and pedestrian tracking since cars and pedestrians appear in large quantity and diversity.
\subsection{Experimental Settings}
\noindent \textbf{Datasets.} KITTI \cite{geiger2012KITTI} and Waymo Open Dataset (WOD) \cite{sun2020scalability} are used for training and validation. Since the labels of the KITTI test set are not provided, following previous works \cite{giancola2019leveraging, qi2020p2b, zheng2021box}, we use the training set to train and evaluate. The training set is spilt as follows: scenes 0-16 for training, scenes 17-18 for validation, and scenes 19-20 for testing. For WOD, we follow \cite{zhou2022pttr} to transform the WOD into class-balanced tracklets that tracking paradigms can handle. 

\noindent \textbf{Evaluation Metrics.} Our evaluation metrics use the One Pass Evaluation (OPE) \cite{wu2013online} from single object tracking. The Success metrics are defined as the Area Under Curve (AUC) of the IoU between the predicted box and the GT. The Precision metrics are defined as the AUC of the distance between the centers for errors from 0 to 2m.

\noindent \textbf{Implementation details.} Following \cite{qi2020p2b}, template number of points sets as $N=512$ and search area sets as $M=1024$ by randomly points discarding and duplicating. We obtain template and search area seeds by three set abstraction layers with query radii of 0.3, 0.5, and 0.7, yielding $M_1=256$, $N_1=128$ and we set the seed feature dimension $d_1=128$. Object-preserved sampling selects highlighted search area seeds with top 128 scores, yielding $M_2=128$, and local feature aggregation generates object-preserved seeds with feature dimension $d_2=128$.

Targeted-feature transformation module uses two 3-layer MLPs with batch normalization (BN) and a ReLU activation layer. For highlighting score prediction, object preserved sampling uses 3-layer MLP and Sigmoid function for object score prediction. In the object localization network, we set the voxel side length $r=0.3$m, the channel of BEV feature map $c=128$, and we generate 4 stacked center maps ($n=4$). For loss regulation, we set $\lambda_1=1.0$,  $\lambda_2=2.0$, and $\lambda_3=0.5$.

\noindent \textbf{Training and testing.} We aggregate the first GTBB of the current tracklet and the previous GTBB as the template, which is applied with a random offset on the previous GTBB to augment data. The search areas are augmented similarly by enlarging the current GT with an offset, for all experiments, this offset sets as 2m. Additionally, we apply the Adam optimizer \cite{kingma2014adam} with an initial learning rate of 0.001 and decreased by 5 times after 10 epochs. We train the OPSNet on a single NVIDIA RTX2080 Ti GPU with a batchsize of 32. OPSNet can obtain satisfying results after about 40 epochs.
\begin{figure*}
    \centering
    \includegraphics[scale=0.30]{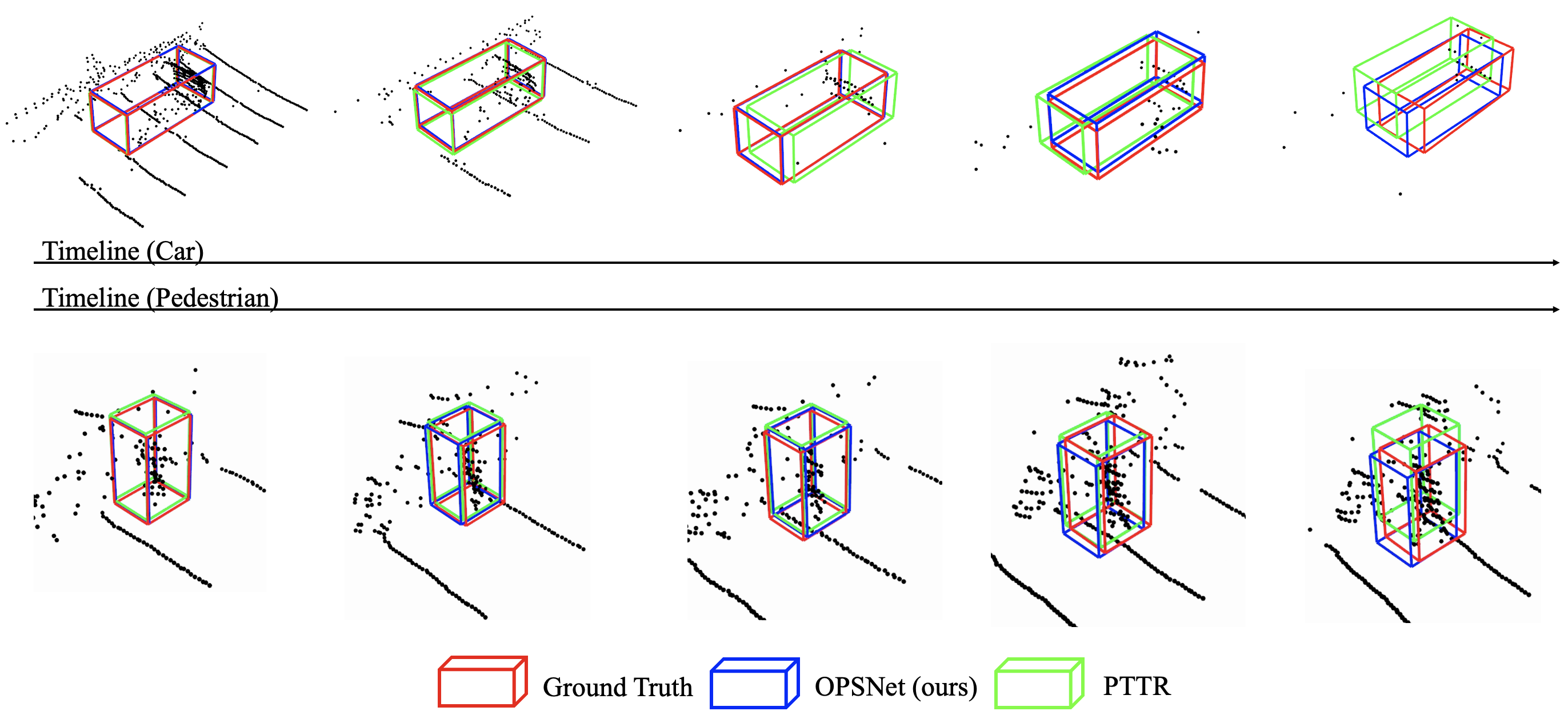}
        \caption{Result comparison visualization with PTTR \cite{zhou2022pttr}. We can observe that the predicted BBoxes by OPSNet hold tight to the ground truths. }
    \label{fig:visual}
\end{figure*}

\begin{table}[]
\centering
\caption{Comparison with other state-of-the-art approaches on the KITTI dataset. The instance frame number of each category is shown under category names, and the mean denotes the average results of all categories. Bold presents the best performance.}
\label{kitti_result}
\resizebox{\columnwidth}{!}{%
    \begin{tabular}{ccccccc} 
    \toprule[2pt]%
    \multicolumn{2}{c|}{Category}             & Car  & Ped  & Van  & Cyclist & Mean  \\
    \multicolumn{2}{c|}{Frame num.}                                      & 6424 & 6088 & 1248 & 308     & 14068 \\ \midrule[1pt]%
    Success   & \multicolumn{1}{c|}{P2B \cite{qi2020p2b}}      & 56.2 & 40.8 & 40.8 & 32.1    & 39.5  \\
                               & \multicolumn{1}{c|}{BAT \cite{zheng2021box}}     & 60.5 & 42.1 & 52.4 & 33.7    & 47.2  \\
                               & \multicolumn{1}{c|}{LTTR \cite{cui20213d}}     & 65.0 & 33.2 & 35.8 & 66.2    & 50.0  \\
                               & \multicolumn{1}{c|}{V2B \cite{hui20213d}}      & 70.5 & 48.3 & 50.1 & 40.8    & 58.4  \\
                               & \multicolumn{1}{c|}{PTTR \cite{zhou2022pttr}}     & 65.2 & 50.9 & 52.5 & 65.1    & 58.4  \\
                               & \multicolumn{1}{c|}{STNet \cite{hui20223d}}    & 72.1 & 49.9 & 58.0 & 73.5    & 61.3  \\
                               & \multicolumn{1}{c|}{M2-Track \cite{zheng2022beyond}} & 65.5 & 61.5 & 53.8 & 73.2    & 62.9  \\ 
                               & \multicolumn{1}{c|}{OPSNet (ours)}     &\textbf{77.0} & \textbf{65.6} & \textbf{77.7} & \textbf{85.5}    & \textbf{72.3}  \\ \midrule[1pt]%
    Precision & \multicolumn{1}{c|}{P2B \cite{qi2020p2b}}      & 72.8 & 49.6 & 48.4 & 44.7    & 53.9  \\
                               & \multicolumn{1}{c|}{BAT \cite{zheng2021box}}      & 77.7 & 70.1 & 67.0 & 45.4    & 65.1  \\
                               & \multicolumn{1}{c|}{LTTR \cite{cui20213d}}     & 77.1 & 56.8 & 45.6 & 89.9    & 67.4  \\
                               & \multicolumn{1}{c|}{V2B \cite{hui20213d}}     & 81.3 & 73.5 & 58.0 & 49.7    & 75.2  \\
                               & \multicolumn{1}{c|}{PTTR \cite{zhou2022pttr}}     & 77.4 & 81.6 & 61.8 & 90.5    & 77.8  \\
                               & \multicolumn{1}{c|}{STNet \cite{hui20223d} }   & 84.0 & 77.2 & 70.6 & \textbf{93.7}    & 80.1  \\
                               & \multicolumn{1}{c|}{M2-Track \cite{zheng2022beyond}} & 80.8 & 88.2 & 70.7 & 93.5    & 83.4  \\ 
                               & \multicolumn{1}{c|}{OPSNet (ours)}     & \textbf{86.1} & \textbf{90.3} & \textbf{85.8} & \textbf{93.7}    & \textbf{88.0}  \\ \bottomrule[2pt] 
    \end{tabular}%
}
\end{table}

\begin{table}[]
\centering
\caption{Comparison with other state-of-the-art approaches on the WOD dataset.}
\label{wod_result}
\resizebox{\columnwidth}{!}{%
\begin{tabular}{lc|cccc}
\toprule[2pt]
                                               & Category               & \multicolumn{4}{c}{Vehicle}                                   \\ \midrule[1pt]%
 &
  \begin{tabular}[c]{@{}c@{}}Level\\ Frame num.\end{tabular} &
  \begin{tabular}[c]{@{}c@{}}Easy\\ 67832\end{tabular} &
  \begin{tabular}[c]{@{}c@{}}Medium\\ 61252\end{tabular} &
  \begin{tabular}[c]{@{}c@{}}Hard\\ 56647\end{tabular} &
  \begin{tabular}[c]{@{}c@{}}Mean\\ 185371\end{tabular} \\ \midrule[1pt]%
  Success                       & BAT \cite{zheng2021box}                    & 61.0          & 53.3          & 48.9          & 54.7          \\
                                               & V2B \cite{hui20213d}                   & 64.5          & 55.1          & 52.0          & 57.6          \\
                                               & STNet \cite{hui20223d}                 & 65.9          & 57.5          & 54.6          & 59.7          \\
                                               & OPSNet (ours)          & \textbf{68.8} & \textbf{58.7} & \textbf{55.2} & \textbf{61.3} \\ \midrule[1pt]%
  Precision                     & BAT \cite{zheng2021box}                   & 68.3          & 60.9          & 57.8          & 62.7          \\
                                               & V2B \cite{hui20213d}                   & 71.5          & 63.2          & 62.0          & 65.9          \\
                                               & STNet \cite{hui20223d}                 & 72.7          & 66.0          & 64.7          & 68.0          \\
                                               & OPSNet (ours)          & \textbf{74.4} & \textbf{66.9} & \textbf{65.1} & \textbf{69.1} \\ \midrule[0.5pt] \midrule[0.5pt]%
\multicolumn{1}{c}{}                           & Category               & \multicolumn{4}{c}{Pedestrian}                                \\ \midrule[1pt]%
\multicolumn{1}{c}{} &
  \begin{tabular}[c]{@{}c@{}}Level\\ Frame num.\end{tabular} &
  \begin{tabular}[c]{@{}c@{}}Easy\\ 85280\end{tabular} &
  \begin{tabular}[c]{@{}c@{}}Medium\\ 82253\end{tabular} &
  \begin{tabular}[c]{@{}c@{}}Hard\\ 74219\end{tabular} &
  \begin{tabular}[c]{@{}c@{}}Mean\\ 241752\end{tabular} \\ \midrule[1pt]%
\multicolumn{1}{c}{Success}   & BAT \cite{zheng2021box}                   & 19.3          & 17.8          & 17.2          & 18.2          \\
\multicolumn{1}{c}{}                           & V2B \cite{hui20213d}                   & 27.9          & 22.5          & 20.1          & 23.7          \\
\multicolumn{1}{c}{}                           & STNet \cite{hui20223d}                 & 29.2          & 24.7          & 22.2          & 25.5          \\
\multicolumn{1}{c}{}                           & OPSNet (ours) & \textbf{32.2} & \textbf{28.4} & \textbf{25.1} & \textbf{28.7} \\ \midrule[1pt]%
\multicolumn{1}{c}{Precision} & BAT \cite{zheng2021box}                   & 32.6          & 29.8          & 28.3          & 30.3          \\
\multicolumn{1}{c}{}                           & V2B  \cite{hui20213d}                  & 43.9          & 36.2          & 33.1          & 37.9          \\
\multicolumn{1}{c}{}                           & STNet \cite{hui20223d}                 & 45.3          & 38.2          & 35.8          & 39.9          \\
\multicolumn{1}{c}{}                           & OPSNet (ours) & \textbf{47.4} & \textbf{41.9} & \textbf{37.6} & \textbf{42.5} \\ \bottomrule[2pt]%
\end{tabular}%
}
\end{table}

\subsection{Results}
\noindent \textbf{Evaluation on KITTI.} As shown in \cref{kitti_result}, our OPSNet shows a large improvement over SOTA methods, All the methods besides M2-track \cite{zheng2022beyond} apply the Siamese-based network. Our OPSNet outperforms the SOTA tracking paradigm M2-track with a large margin of $\sim$9.4\% / $\sim$4.6\% on average success / precision and the SOTA PTTR \cite{zhou2022pttr} based on PointNet++ of $\sim$13.9\% / $\sim$10.2\%. Our OPSNet achieves the highest tracking performance in all four categories. As shown in \cref{fig:visual}, we compare the proposed OPSNet against PTTR over the car and pedestrian sequences since PTTR adopts relation-aware sampling for object preservation. The error gradually accumulates in PTTR tracking since the bad template misleads the search area sampling, but the object-preserved search area seeds generated by object-preserved sampling assist the OPSNet to perform robust tracking.

\noindent \textbf{Evaluation on WOD.} To validate the generalization ability of our OPSNet, we use our OPSNet that pre-trained over the KITTI dataset for testing tracking on Waymo Open Dataset (WOD) following \cite{hui20213d}. According to the point number of the first frame, WOD can be divided into three subsets of different tracking levels of difficulty, including easy, medium, and hard. Even though tracking frames are far more than KITTI has, our OPSNet still shows the generalization ability of the tracking performance on WOD, which outperforms the baseline BAT \cite{zheng2021box} with a gain of $\sim$6.6\% / $\sim$6.4\% on the category Vehicle, results as \cref{wod_result} shows. 

\begin{figure}
    \centering
    \includegraphics[scale=0.32]{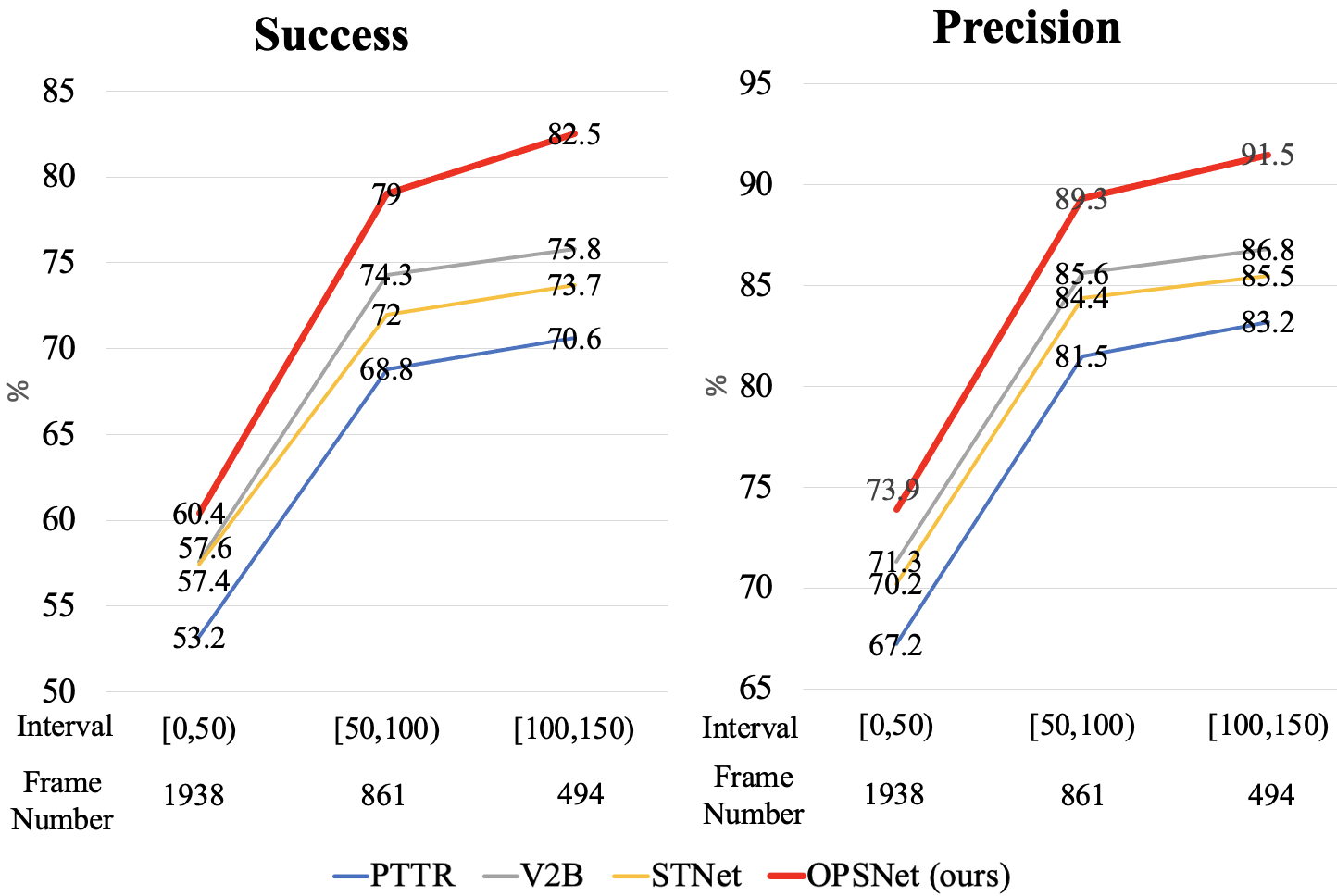}
        \caption{Evaluated mean success / precision of the three extreme sparse point number intervals on the KITTI Car category.}
    \label{fig:number_intervals}
\end{figure}

\noindent \textbf{Tracking on sparse point clouds.} Handling extreme sparse scenes is a great challenge in 3D SOT tasks. We test our OPSNet on the object of extreme sparse points, which is divided into three object point number intervals including $[0,50),[50,100),[100,150)$. We compare the tracking result of our OPSNet, PTTR \cite{zhou2022pttr}, V2B\cite{hui20213d}, and STNet\cite{hui20223d}. As \cref{fig:number_intervals} shows, our OPSNet achieves the highest tracking performance on three intervals with $\sim$2.8\% / $\sim$2.6\%, $\sim$4.7\% / $\sim$3.7\%, $\sim$6.7\% / $\sim$4.7\% success / precision gain over the V2B. Experiment results demonstrate that our OPSNet effectively preserves sparse objects and handles extreme sparse point clouds well.

\begin{figure}
    \centering
    \includegraphics[scale=0.34]{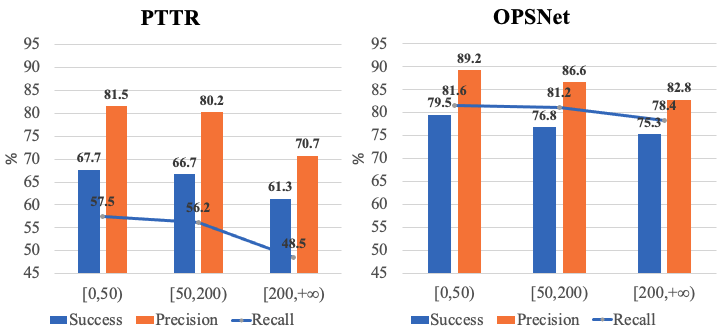}
        \caption{Evaluated mean success / precision and recall rate of the three sequence length intervals on the KITTI Car category.}
    \label{fig:length_intervals}
\end{figure}

\noindent \textbf{Tracking on long sequences.} Siamese-based methods probably lose the object when tracking on long sequences. Therefore the performance on long sequences evaluates the robustness of SOT methods. PTTR adopts their relation-aware sampling for object preservation, therefore we test our OPSNet and PTTR on sequence length intervals $[0,50),[50,200),[200,+\infty)$ and compute average recall rate after sampling as the indicator of object preservation.
As \cref{fig:length_intervals} shows, our OPSNet outperforms the PTTR on three intervals with $\sim$11.8\% / $\sim$7.7\%, $\sim$10.1\% / $\sim$6.4\%, $\sim$14.0\% / $\sim$12.1\% success / precision gain and $\sim$22.0\%, $\sim$20.6\%, $\sim$29.9\% recall rate gain.

\noindent \textbf{Running speed.} We evaluate the inference speed of our model on a single RTX2080 Ti GPU. Our method achieves 23 FPS, including 21 ms for pre-processing point clouds, 21 ms for network forward propagation, and 2 ms for post-processing. We also test V2B, and STNet in the default settings run with 20 FPS and 19 FPS, respectively. 

\subsection{Ablation Studies}
We conduct ablation studies to evaluate the modules proposed in OPSNet, the results are shown with OPE evaluation metrics and based on the KITTI dataset.

\begin{figure*}
    \centering
    \includegraphics[scale=0.38]{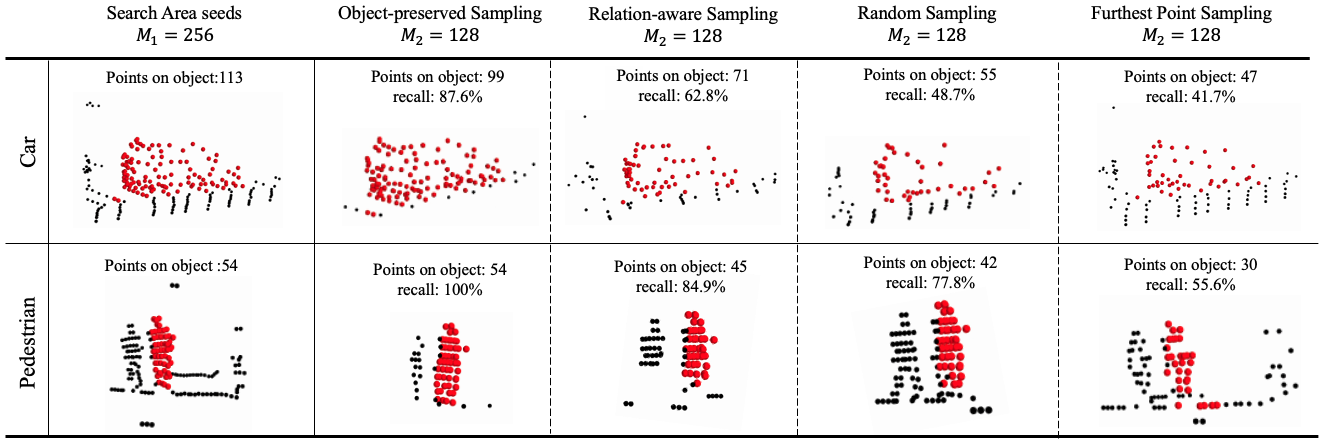}
        \caption{Comparison on sampling strategies: our object-preserved sampling (OPS), relation-aware sampling (RAS) proposed in PTTR \cite{zhou2022pttr}, random sampling (RS) employed in \cite{qi2020p2b}, and furthest points sampling (FPS) proposed in PointNet++ \cite{qi2017pointnet++}. Note that red points denote the object and black ones denote the background.}
    \label{fig:visual_comparison}
\end{figure*}

\noindent \textbf{Comparison on sampling strategies.} We compare our proposed object preserved sampling (OPS) with existing unsupervised sampling strategies including relation-aware sampling (RAS) \cite{zhou2022pttr}, random sampling (RS) \cite{qi2020p2b}, and furthest point sampling (FPS) \cite{qi2017pointnet++} on sampling 128 candidates from the 256 search area seeds, as \cref{fig:visual_comparison} shows. We compute the recall rates for object points after sampling which demonstrate the effectiveness of our OPS. 
We also replace our OPS with RAS, RS, and FPS in OPSNet, default OPSNet outperforms clear success / precision gain of $\sim$4.7\% / $\sim$3.6\% on the car category and $\sim$9.0\% / $\sim$8.8\% on the pedestrian over the FPS baseline, comparison shown in \cref{sample_compare}.  

\noindent \textbf{Ablation studies on object highlighting module.} We conduct ablation experiments to validate the effectiveness of the object highlighting components, respectively, as \cref{components_compare} shows. We report the results of cars and pedestrians tracking on the KITTI dataset. Only a single cross-correlation approach cannot perform the best tracking ($\sim$6.9\% / $\sim$3.5\% decrease with only consistent cross-correlation and $\sim$6.4\% / $\sim$3.0\% decrease with only discrepant cross-correlation on Car). For analyzing feature-targeted transformation, we simply divide the search area into two subsets without utilizing MLPs, so that the cross-correlation cannot be adaptively conducted ($\sim$7.7\% / $\sim$3.6\% decrease without the feature-targeted transformation on Car). Results demonstrate that object augmentation further improves tracking performance ($\sim$2.5\% / $\sim$1.8\% gain on the category Car). 

\begin{table}[]
\centering
\caption{Tracking performance comparison on different sampling approaches. Shown by the evaluation metrics success / precision on the category car and pedestrian.}
\label{sample_compare}
\begin{tabular}{c|cc}
\toprule[2pt]%
Method     & Car         & Pedestrian   \\ \midrule
RS \cite{qi2020p2b}     & 72.9 / 83.7 & 58.2 / 82.1  \\
FPS \cite{qi2017pointnet++}     & 71.2 / 81.8 & 52.4 / 80.5  \\
RAS \cite{zhou2022pttr}       & 73.3 / 83.0 & 61.0 / 84.5  \\ \midrule
OPS (ours) & \textbf{77.0} / \textbf{86.1} & \textbf{65.6} / \textbf{90.3}  \\ \bottomrule[2pt]%
\end{tabular}%

\end{table}

\begin{table}[]
\centering
\caption{Ablation studies results on object highlighting module.}
\label{components_compare}
\resizebox{\columnwidth}{!}{%
\begin{tabular}{c|cc}
\toprule[2pt]%
                                     & Car                  & Pedestrian           \\ \midrule[1pt]%
\multicolumn{1}{c|}{w/o object augmentation}        & 74.5 / 84.3          & 62.2 / 86.9          \\
\multicolumn{1}{c|}{w/o feature-targeted transformation}        & 69.3 / 82.5          & 59.1 / 79.8          \\
\multicolumn{1}{c|}{only consistent cross-correlation}      & 70.1 / 82.6          & 59.9 / 80.1          \\
\multicolumn{1}{c|}{only discrepant cross-correlation}      & 70.6 / 83.1          & 60.3 / 80.5          \\
\midrule[1pt]%
\multicolumn{1}{c|}{default setting}       & \textbf{77.0 / 86.1} & \textbf{65.6 / 90.3} \\ 
\bottomrule[2pt]%
\end{tabular}
}
\end{table}
\begin{table}[]
\centering
\caption{Tracking performance comparison on common object locating approaches.}
\label{locating_compare}
\begin{tabular}{c|cc}
\toprule[2pt]%
                                     & Car                  & Pedestrian           \\ \midrule[1pt]%
\multicolumn{1}{c|}{RPN \cite{qi2020p2b,zheng2021box}}        & 68.0 / 79.6          & 48.8 / 72.5          \\
\multicolumn{1}{c|}{C2R \cite{zhou2022pttr}}        & 71.5 / 83.2          & 57.1 / 80.2          \\
\midrule[1pt]%
\multicolumn{1}{c|}{OLN (ours)}       & \textbf{77.0 / 86.1} & \textbf{65.6 / 90.3} \\ 
\bottomrule[2pt]%
\end{tabular}
\end{table}

\noindent \textbf{Ablation studies on object localization network.} We replace our object localization network (OLN) with the VoteNet-based approaches including Region Proposal Network (RPN) applied in \cite{qi2020p2b,zheng2021box} and coarse-to-refine regression (C2R) proposed in \cite{zhou2022pttr} to validate the tracking performance. VoteNet regresses the offset from point to center with a score as proposals. However, when facing sparse scenes VoteNet is hard to generate high-quality proposals. Our object localization network does not require generating numerous 3D proposals thus it can handle the above issue and outperforms baseline RPN with $\sim$9.0\% / $\sim$6.5\% and $\sim$16.8\% / $\sim$17.8\% increase on the category Car and Pedestrian.  

\section{Conclusion} In this paper, we propose a novel Object Preserving Siamese Network (OPSNet) for single object tracking on point clouds. Our OPSNet aims to enhance the object-aware features and extract discriminative features from template and search area with the object highlighting module, maintain object integrity and drop redundant background points with the object-preserved sampling, and predict accurate BBoxes with the object localization network.

The experimental results ($\sim$9.4\% and $\sim$2.5\% success gain on KITTI and WOD respectively) demonstrate that our OPSNet achieves high-performance object tracking and possesses generalization ability, indicate that object highlighting module and object-preserved sampling effectively handle sparse point clouds and assist the object localization network to perform robust tracking on long sequences.


{\small
\bibliographystyle{ieee_fullname}
\bibliography{egbib}
}

\end{document}